# PeSOTIF: a Challenging Visual Dataset for Perception SOTIF Problems in Long-tail Traffic Scenarios

Liang Peng, Jun Li, Wenbo Shao, and Hong Wang ✉, *Member, IEEE*

*Abstract*—Perception algorithms in autonomous driving systems confront great challenges in long-tail traffic scenarios, where the problems of Safety of the Intended Functionality (SOTIF) could be triggered by the algorithm performance insufficiencies and dynamic operational environment. However, such scenarios are not systematically included in current open-source datasets, and this paper fills the gap accordingly. Based on the analysis and enumeration of trigger conditions, a high-quality diverse dataset is released, including various long-tail traffic scenarios collected from multiple resources. Considering the development of probabilistic object detection (POD), this dataset marks trigger sources that may cause perception SOTIF problems in the scenarios as key objects. In addition, an evaluation protocol is suggested to verify the effectiveness of POD algorithms in identifying the key objects via uncertainty. The dataset never stops expanding, and the first batch of open-source data includes 1126 frames with an average of 2.27 key objects and 2.47 normal objects in each frame. To demonstrate how to use this dataset for SOTIF research, this paper further quantifies the perception SOTIF entropy to confirm whether a scenario is unknown and unsafe for a perception system. The experimental results show that the quantified entropy can effectively and efficiently reflect the failure of the perception algorithm.

## I. INTRODUCTION

With the rapid development of deep learning and computing environment, artificial intelligence technologies have significantly impacted autonomous driving systems, especially perception subsystems. However, deep learning-based algorithms in perception, prediction, and planning subsystems typically provide black box solutions, the randomness and uncertainties can induce safety risks to autonomous vehicles. These types of risks are within the scope of the safety of the intended functionality (SOTIF), which refers to the absence of unreasonable risk due to hazards caused by performance limitation, functional insufficiency, or reasonably foreseeable misuse [1].

There are two necessary conditions for the SOTIF problems: performance limitation or functional insufficiency and trigger conditions. For a deep learning-based perception algorithm, the former condition is relevant to the amount of training data, and the performance commonly degrades when confronted with long-tail scenarios. Research on detection of unknown categories and out-of-distribution objects is trying to alleviate this problem. The latter condition relates to the operational design domain (ODD), including the environment-related conditions and object-related conditions.

Liang Peng, Jun Li, Wenbo Shao, and Hong Wang are with State Key Laboratory of Automotive Safety and Energy, School of Vehicle and Mobility in Tsinghua University, Beijing, China (e-mail: pengl-l20@mails.tsinghua.edu.cn; lijun1958@tsinghua.edu.cn; swb19@mails.tsinghua.edu.cn; and hong_wang@mail.tsinghua.edu.cn). Hong Wang is the corresponding author.

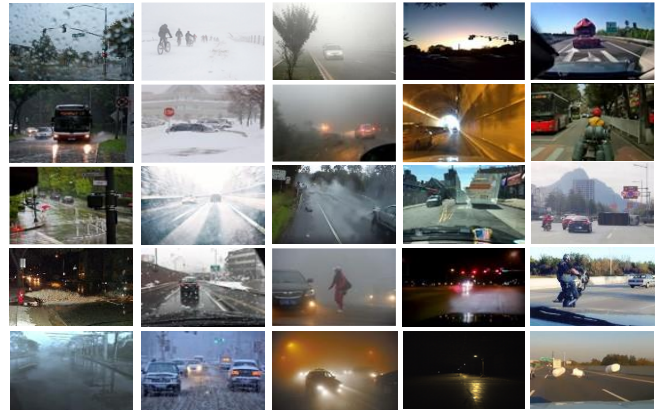

(a) Rain  (b) Snow  (c) Particulate  (d) Illumination  (e) Object

Fig. 1: A brief view of the PeSOTIF dataset. Columns (a)(b)(c)(d) list some sample images of the 'environment' subset, and column (e) shows some sample images of the 'object' subset. The environmental factors can be natural or artificial and have different effects. For example, direct sunlight comes from the sky and improves the overall illumination intensity of the surrounding environment, while full-beam headlamps are moveable and highly directional light sources.

The weather conditions, such as rainy, foggy, and snowy weather, and the lighting conditions, such as direct sunlight and full-beam headlamps, belong to the environment-related trigger conditions which can significantly degrade the perception ability. In contrast, temporary obstacles (*e.g.*, landslips and traffic cones), disturbing objects (*e.g.*, animals and falling tires from surrounding vehicles), and common road users with unexpected appearances and postures belong to the object-related trigger conditions which can degrade the cognition ability and thereby lead to perception SOTIF risks. The combination of different trigger conditions can form various perception SOTIF scenarios. During the collection of existing data sets, the above scenarios occasionally occur. However, the dataset proposed in this paper aims at these trigger conditions and extracts diverse long-tail traffic scenarios from various data sources, as shown in Fig. 1.

Many approaches have emerged for uncertainty estimation when studying the black box characteristics and performance limitations of deep learning-based algorithms. Among them, sampling-based methods have been the most popular for algorithms of autonomous driving perception subsystems, such as semantic segmentation and object detection, due to their simplicity and convenience. However, most of the related studies have focused on improving the evaluation metrics based on the estimated uncertainty, without further considering the performance boundary and safety of the algorithm.

In this paper, a challenging visual dataset PeSOTIF for 2D object detection and probabilistic object detection (POD) in long-tail traffic scenarios is proposed, aiming to promote the research on perception SOTIF problems. The first version of the dataset is published under CC BY-NC-SA 4.0 license and is available at: https://github.com/SOTIF-AVLab/PeSOTIF. The main contributions are as follows:

- This paper releases the first diverse dataset that is systematically collected and organized based on the analysis of trigger conditions for studying perception SOTIF problems in long-tail traffic scenarios.

- About 2.27 key objects in each scenario that may trigger perception SOTIF problems are manually marked as hard objects in annotations of both YOLO and COCO formats.

- This paper suggests an evaluation protocol for verifying the effectiveness of perception uncertainty estimated by the POD algorithms.

- A demonstration of testing the performance of a POD algorithm on PeSOTIF is proposed, where the solution can quantify the perception uncertainty online and propagate it to the downstream modules.

## II. RELATED WORKS

Significant studies have been devoted to establishing critical traffic scenarios and estimating and evaluating the epistemic uncertainty of deep learning-based perception algorithms.

### A. Perception Data Sets for Vision-Based Object Detection

The early visual data sets, such as the widely used COCO dataset, are constructed for the computer vision field and include many indoor scenarios [2]. To study the performance limitation and improvement methods of the vision-based perception algorithms, synthetic datasets, such as the Raindrop dataset [3], and specific datasets collecting scenarios under severe environmental conditions, such as the ExDark dataset [4], have been released. Among all these data sets, there are some but not many traffic scenarios, not to mention the SOTIF-related long-tail scenarios.

As vision-based perception algorithms are adopted in autonomous driving systems, an increasing number of traffic datasets for on-board perception tasks have been introduced [5]. The KITTI dataset has been widely used as the benchmark for perception tasks [6]. The nuScenes dataset containing 40k labeled data is becoming a newly developing benchmark for algorithm evaluation and competition [7]. Besides, UC Berkeley released their large-scale BDD100k dataset containing about 100k frames and 2M labeled data [8]. In addition, vehicle industry is also contributing. Waymo released their dataset with 12M labeled data [9]. Audi and Honda also released their A2D2 [10] and H3D [11] data sets. Although there have been some specific datasets introduced in recent years, such as the Radiate dataset that includes several traffic scenarios under rainy and snowy weather [12], few datasets have systematically involved the SOTIF-related trigger conditions [13].

### B. Uncertainty Estimation Methods for POD Algorithms

Uncertainty estimation is one of the ways to study the algorithm performance under SOTIF-related scenarios. Perception uncertainties were decomposed into seven main sources by Czarnecki *et al* [14]. Meanwhile, prediction uncertainty of deep neural networks was divided into epistemic and aleatoric uncertainties by Kendall *et al.* [15]. The aleatoric uncertainty reflects the influence of input noise and cannot be eliminated. However, the epistemic uncertainty of a deep learning-based perception algorithm reflects its perception and cognition abilities and can be reduced by increasing the amount of training data.

Research on epistemic uncertainty estimation of deep neural networks includes three main types of methods: (1) Bayesian-based methods, such as stochastic variational inference (SVI), which provide precise theoretical derivations and use fewer assumptions, but typically have high computational complexity [16]; (2) sampling-based methods, such as deep ensembles (DE), which estimate the parameters of the posterior distribution through easily obtained samples and have the potential for real-time applications [17]; (3) other methods (*i.e.*, non-Bayesian and non-sampling methods), such as deep evidential regression (DER), which can obtain the aleatoric uncertainty that reflects the data noise in addition to the epistemic uncertainty, but require more modifications to the network structures and the related theories and mechanisms are complicated [18].

### C. Applications of Uncertainty for SOTIF Research

The studies utilizing the uncertainty estimation methods can be roughly divided into three main aspects. Firstly, many studies take advantage of the redundancy and complementarity of sampling-based methods, such as deep ensembles, to improve the evaluation indicators, such as mean Average Precision (mAP), of probabilistic object detectors [19]. Then, some metrics which consider both the accuracy and uncertainty of an algorithm, such as probabilistic detection quality (PDQ), are proposed to obtain comprehensive evaluation results [20]. In addition, some studies have designed selection and rejection mechanisms based on the uncertainty and proposed corresponding evaluation metrics, such as variation ratio (VR) [21]. However, such simple mechanisms are not sufficient for safety-critical applications, such as autonomous driving, and the VR cannot reflect missing objects and ghost detections.

Therefore, it is recommended in this paper to quantify the perceptual risks through uncertainty estimation and propagate the obtained information to the downstream subsystems for mitigation. This work aims to construct a test dataset for perception SOTIF problems and design an evaluation protocol to verify the effectiveness of the above approach.

## III. PESOTIF DATASET

### A. Analysis of Trigger Conditions

The research on SOTIF risk basically starts from the analysis of trigger conditions and trigger sources in the dynamic operating environment. Huang *et al.* proposed an analysis framework of perceptual trigger conditions according to the chain of events model [22], which provides reference for this paper. Besides, this work also theoretically considers the physical and algorithmic principles of perception systems.

Based on the above analysis, this paper lists the trigger conditions of the autonomous perception system, and divides the trigger sources into three levels. Meanwhile, in terms of the analysis of the endless new data, the list has been constantly updating and improving. Table I shows some trigger conditions related to the visual sensors in the list.

TABLE. I: Partial Trigger Conditions of Visual Sensors

| Primary Label | Secondary Label | Tertiary Label | Trigger Sources |
|---|---|---|---|
| Environment | Rain | Natural | stains, streaks, haze, splashes, puddles, etc. |
| | | Handcraft | hydrants, sprinklers, synthetic raindrops, etc. |
| | Snow | Natural | stains, flakes, haze, snow covers, etc. |
| | | Handcraft | snowplows, synthetic snowflakes, etc. |
| | Particulate | Natural | scattering, haze, Tyndall effect, etc. |
| | | Handcraft | vehicle emissions, crash sites, construction dust, etc. |
| | Illumination | Natural | direct sunlight, reflections, shadows, etc. |
| | | Handcraft | full-beam headlamps, unusual color temperatures, etc. |
| Object | Common | Appearance | overlaps, occlusions, graffiti, customized cars, etc. |
| | | Posture | overturned cars, people lying on the ground, stunts, etc. |
| | Uncommon | | animals, landslips, falling objects, etc. |

According to Table I, the PeSOTIF dataset is mainly divided into two subsets, in which the 'environment' subset includes scenarios that degrade the perception ability, while the 'object' subset includes scenarios that degrade the cognition ability. In the 'environment' subset, the images are divided into 'rain', 'snow', 'particulate', and 'illumination' subsets, and each is further divided into 'natural' and 'handcraft' subsets. For example, the direct sunlight is a natural factor while the full-beam headlamp is a handcraft factor in the 'illumination' subset. In the 'object' subset, scenarios containing common road users with unusual appearances and postures are put into the 'common' subset. In addition, scenarios containing temporary obstacles and disturbing objects are put into the 'uncommon' subset.

*B. Data Collection Approaches*

The PeSOTIF dataset aims to be a diverse test dataset for perception SOTIF problems currently, thus collects key frames of critical traffic scenarios extracted from multiple data sources. In the future, after collecting sufficient long-tail scenarios, decomposing a training dataset to improve the algorithm performance will be considered.

The first part of the data comes from experiments designed to study trigger conditions and performance boundaries under different rain intensity and illumination intensity in the previous work [23]. A FLIR color camera was used to collect the data. Model: GS3-U3-41C6C-C; CMOSISCMV4000-3E5; global shutter; gain: -7.742dB to 24dB; high dynamic range: cycle 4 gain and exposure presets; frequency: 90Hz; 4.1MP; image size: 2048×2048; lens: FA1215A; lens hFov: -53.6°-53.6°; lens vFov: -41.8°-41.8°; lens focal length: 12mm. About two frames in each group of experiments were extracted and added to the PeSOTIF dataset.

The second part of the data comes from traffic accident videos. There are thousands of videos uploaded by the perception task group of the China SOTIF technical alliance. These videos were screened from road test data, public traffic accident databases, and even clips-sharing sites like YouTube. This work has browsed these videos, further filtered out the scenarios related to perception SOTIF problems. Then, the key frames that may cause perception failure and traffic accidents were located and added to the PeSOTIF dataset.

In addition, this paper also selected some images that meet our requirements from the test subset of the existing data sets. For on-board traffic data sets, some scenarios affected by the perceptual trigger conditions were added to the PeSOITF dataset. For visual detection data sets under adverse environment, some outdoor data was extracted.

*C. Ground Truth Annotations*

The first version of the PeSOTIF dataset includes 1126 frames of data that covers different weather, seasons, and times of the day. This work labeled 11 categories of traffic participants manually by three experienced human drivers with the tool provided in LabelImg. The categories include car, bus, truck, train, bike, motor, person, rider, traffic sign, traffic light, and traffic cone. In the YOLO format, each frame of image has a .txt file to store its annotations. Besides, the dataset is also reorganized as the COCO format for quick use.

In the studies on POD algorithms, evaluation metrics have been generally designed according to confidence and uncertainty [24], [25]. In addition, it has been assumed that high and low uncertainty levels are associated with the wrong and correct classification results, respectively. However, this work reflects on Uber's fatal accident [26] that the continuous change in perception results confronted with the SOTIF scenarios also indicates its high uncertainty. Although an algorithm may classify an object correctly in a certain frame, the results can change to wrong soon in the following frames. Therefore, when conducting object detection for a single frame extracted from a scenario, outputting high uncertainty during correct classification is meaningful in certain cases. Nonetheless, the uncertainty should be low in most cases of correct classification, but it is difficult to automatically judge a specific situation from a single frame.

Therefore, when labeling the PeSOTIF dataset, in addition to the category and bounding box parameters, a binary factor $f_h$ has been added to the annotation of each object. Factor $f_h$ denotes the subjective evaluation result of the difficulty of an object by human drivers according to the context, and $f_h = 1$ indicates that the object is one key object in the keyframe and the ability to recognize the object may be reduced due to the trigger conditions in the scenario. In the first batch of data, there are 2555 key objects and 2778 key objects labeled in total. In other words, an average of approximately 2.27 key objects and 2.47 normal objects are annotated in each key frame.

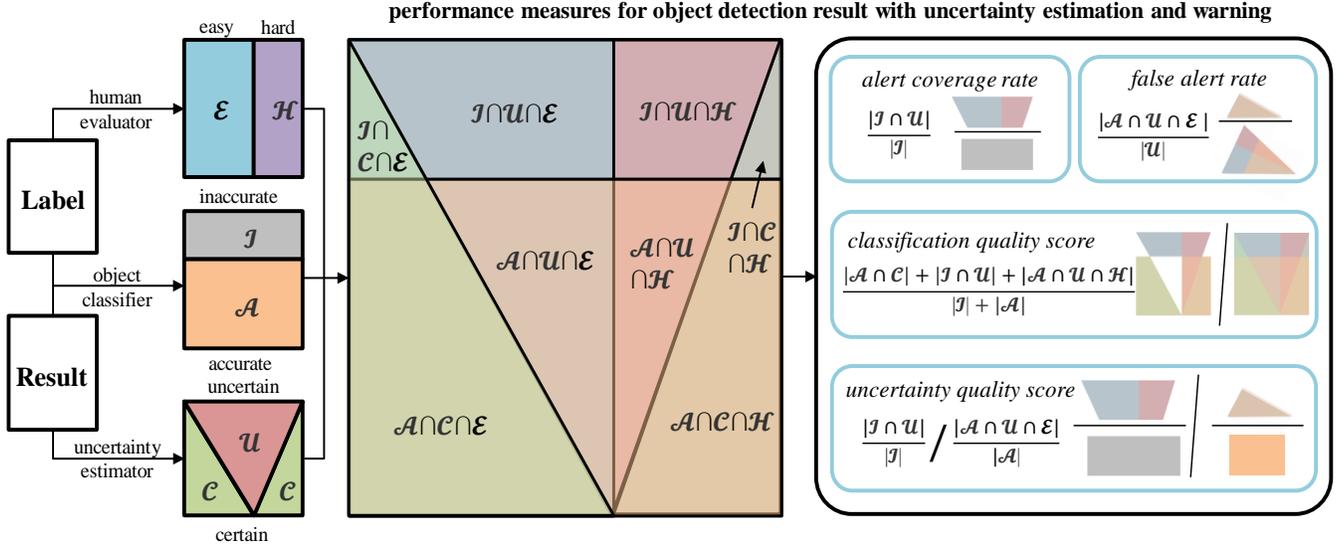

Fig. 2: The evaluation protocol for the PeSOTIF dataset, including the analysis framework and evaluation metrics.

*D. Suggested Evaluation Protocol*

In probabilistic object detectors utilizing sampling-based uncertainty estimation methods, the mean and variance of samples have often been used to approximate the parameters of the posterior distribution [27]. Meanwhile, Shannon entropy has also been commonly adopted to handle the class uncertainty of a single-label classification [28]. Therefore, this work quantifies the perception SOTIF entropy of multi-label object detectors as follows:

$$p_c = p(y=c \mid x, D) \approx \frac{1}{T}\sum_{t=1}^{T} p(y=c \mid x, W_t) \quad (1)$$

$$E^* = -\sum_{p_c}^{C} [p_c \log p_c + (1-p_c)\log(1-p_c)] \quad (2)$$

$$H = H^* \times [1 + f_p \times (T - d)] \quad (3)$$

where $x$ denotes the input data, $y$ is the output label, $D$ is the training dataset, $C$ is the number of total categories, $c$ represents the identifier of a category, $p_c$ refers to the output probability of the $c$-th category, $T$ is the total number of samples, $t$ represents the identifier of a sample, $W_t$ refers to the weights of the $t$-th sampling model, $H^*$ is the sum of Shannon entropy values of the binary classification results $[p_c, 1-p_c]$ of multiple independent single-label object detection, $H$ refers to the final perceptual SOTIF entropy, $d$ is the number of sampling models which detect the object, and $f_p$ refers to the additional penalty factor for missing objects and ghost detections. $H$ should be propagated to the downstream subsystems, and warnings will be generated to promote safer planning when $H$ exceeds the threshold $\theta_w$.

For the probabilistic object detectors with the functions of uncertainty estimation, this work suggests an evaluation protocol for measuring the performance of the entropy-based warning mechanism, as shown in Fig. 2. The analysis is performed in three dimensions. First, the objects are classified as easy or hard in terms of the human evaluation factor values in their annotations. Second, the results are judged as accurate or inaccurate based on their accordance with the ground-truth data. Third, the results are classified into certain or uncertain bins according to their entropy values and the threshold $\theta_w$. Based on this scheme, four evaluation indicators are introduced in the protocol, and they are defined as follows:

- Alert coverage rate (ACR): the ratio of the number of objects that have been warned to that of objects that should be warned; the higher the ratio is, the better the performance will be;

- False alert rate (FAR): the ratio of the number of objects that have been warned but should not be warned to that of all warned objects; the lower the ratio is, the better the performance will be;

- Classification quality score (CQS): the results consistent with the meaning of classification dimension and uncertainty dimension; the larger the score is, the better the performance will be;

- Uncertainty quality score (UQS): has the similar meaning as CQS; the higher the proportion of high-uncertainty objects in inaccurate results is, and the lower the proportion of high-uncertainty objects in accurate results is, the larger the UQS value will be, and the better the performance will be.

## IV. EXPERIMENTS

Object detection is one of the most challenging perception tasks in autonomous driving for a long time. In recent years, many researches have built probabilistic object detectors by introducing uncertainty estimation into the detector head [29], [30]. Both kinds of algorithms are evaluated in this paper.

*A. 2D Object Detection*

The monocular 2D object detection algorithms can be divided into two categories [31]. Two-stage methods have better accuracy but insufficient efficiency. From R-CNN [32] to Faster R-CNN [33], the researchers strived to reduce the running time while ensuring the accuracy. On the contrary, one-stage methods have high efficiency but lacks accuracy. YOLO series algorithms are developed to improve accuracy while retaining real-time performance [34].

This paper tests four algorithms as baselines for the 2D object detection task, namely YOLOv5 [35], Faster R-CNN [33], RetinaNet [36], and Sparse R-CNN [37]. In order to enable these object detectors to identify 11 categories of traffic participants in the PeSOTIF dataset, this paper uses the BDD100k train dataset (70k images) and a traffic cone dataset (0.27k images) for transfer learning [38]. As the BDD100k dataset is much bigger, the networks are pre-trained 100 epochs with the traffic cone dataset before 5 epochs of training with it.

TABLE. II: Common Baseline Evaluation Results for 2D Detection in the BDD100k Validation Dataset

| Metrics | $mAP_{50}$ | $mAR_{50}$ | $mmAP_{50:95}$ |
|---|---|---|---|
| Faster R-CNN | 0.427 | 0.313 | 0.227 |
| RetinaNet | 0.411 | 0.339 | 0.229 |
| Sparse R-CNN | 0.417 | 0.356 | 0.230 |
| YOLOv5 | 0.415 | 0.321 | 0.224 |
| YOLOv5_60e | 0.712 | 0.473 | 0.516 |

In the training and validating procedures, the confidence threshold is set to 0.001 and the IoU threshold is set to 0.6 in the non-maximum-suppression (NMS) process. Because of the complexity of the BDD100k dataset and the short training schedule, the indicators are lower than other benchmarks. The results are shown in Table II. Then the performance of the baselines on the PeSOITF dataset is evaluated in the testing procedure, where the confidence threshold is set to 0.25 and the IoU threshold is set to 0.45 in NMS. The indicators are much lower due to the high complexity and diversity of the PeSOTIF dataset. The results are shown in Table III.

TABLE. III: Common Baseline Evaluation Results for 2D Detection in the PeSOTIF Test Dataset

| Metrics | $mAP_{50}$ | $mAR_{50}$ | $mmAP_{50:95}$ |
|---|---|---|---|
| Faster R-CNN | 0.311 | 0.273 | 0.163 |
| RetinaNet | 0.353 | 0.335 | 0.193 |
| Sparse R-CNN | 0.284 | 0.325 | 0.153 |
| YOLOv5 | 0.340 | 0.202 | 0.164 |
| YOLOv5_60e | 0.459 | 0.276 | 0.205 |

### B. Probabilistic Object Detection

To come to the general approach, this work focuses on dealing with one specific perceptual SOTIF-related long-tail scenario, namely, the sanitation worker-traffic cone scenario. In this scenario, due to the high similarity between the colors and stripes of sanitation workers' uniforms and traffic cones, object detectors may be misled under the interference of trigger conditions. Specifically, when sanitation workers stoop or sit on the ground, algorithms are more likely to misidentify them, as shown in Fig. 3. In order to obtain better performance, this work continue to train YOLOv5 object detectors until 60 epochs and its performance exceeds ResNet whose $mmAP_{50:95}$ is 0.499 [39], as shown in Table IV.

The deep ensembles method is utilized to estimate the epistemic uncertainty of YOLOv5 object detectors. Then, the pre-NMS techniques and basic sequential algorithm scheme with intra-sample exclusivity (BSASexcl) based on winning label (WL) and intersection over union (IoU) are used to handle the output probabilities according to (1) [40]. Finally, the entropy $H$ is quantified according to (2) and (3).

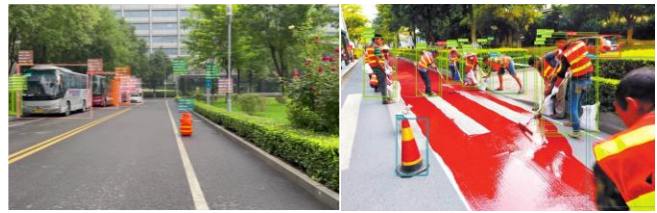

(a) Sanitation worker  (b) Traffic cone

Fig. 3: Characteristic similarity between sanitation workers and traffic cones. A tester sitting on the ground with uniforms in (a) seems very similar to the traffic cone in (b).

The main experimental parameters are set as $C=11$, $T=5$, and $f_p=0.1$. Some sample images in the PeSOTIF dataset detected by the ensemble are shown in Fig. 4. Then the basic evaluation metrics of a sample network and the ensemble of YOLOv5 detectors are given in Table IV. By comparing the evaluation results in Tables II, III, and IV, the performance of the trained networks on the PeSOTIF dataset is much lower than on the BDD test set, which indicates that the traffic scenarios in the PeSOTIF dataset are quite critical despite the out-of-distribution (OOD) problems. In addition, the performance of the ensemble is better than that of the single sampling model.

TABLE. IV: Common Evaluation Results of the Ensemble of Enhanced YOLOv5 Object Detectors

| Metrics | $mAP_{50}$ | $mAR_{50}$ | $mmAP_{50:95}$ |
|---|---|---|---|
| BDD test-Sample_60e | 0.656 | 0.468 | 0.445 |
| BDD test-Ensemble_60e | 0.643 | 0.476 | 0.457 |
| PeSOTIF test-Sample_60e | 0.459 | 0.276 | 0.205 |
| PeSOTIF test-Ensemble_60e | 0.447 | 0.287 | 0.213 |

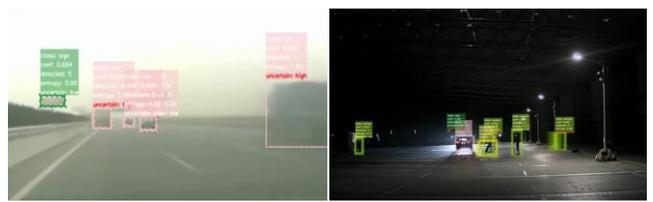

(a) Environment-Particulate-Natural  (b) Environment-Illumination-Handcraft

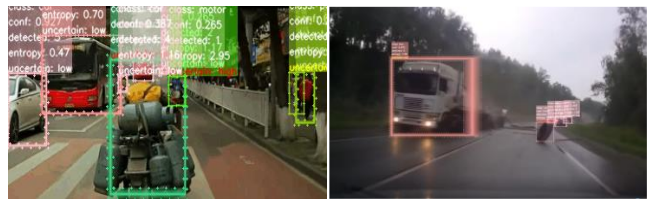

(c) Object-Common-Appearance  (d) Object-Uncommon

Fig. 4: Sample results of the PeSOTIF dataset detected by the ensemble, where objects with high uncertainty are labeled red at the bottom of their information.

The proposed evaluation protocol is used to verify the performance of uncertainty estimation and PeSOTIF-based warning mechanism on the PeSOTIF dataset in this work. The experimental results are presented in Fig. 5. When the graduation value is set as 0.1, with the decrease of the uncertainty threshold, the ACR and CQS increase, the FAR first decreases and then increases, and the UQS first increases and then decreases. Under the uncertainty threshold of 1.2, the minimum FAR is 9.4% and the maximum UQS is 5.010. This would be an optimal solution if pursuing the minimum errors of warning. However, in this case, the ACR is only 53.7%, so many dangerous objects could not be identified. To achieve the best comprehensive performance in practical application, the uncertainty threshold should be set to $\theta_w = 1.0$, where the ACR would increase to 90.0%, the CQS would increase to 0.858, the FAR would increase to 10.8% as a cost, and the UQS would decrease to 4.238.

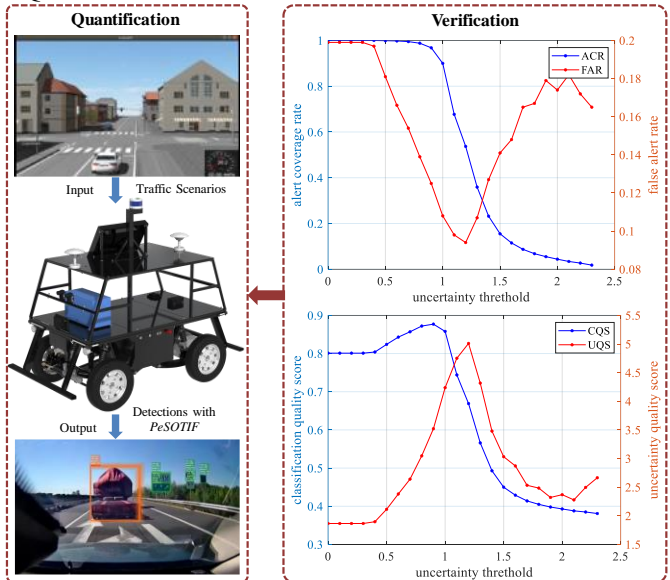

Fig. 5: Quantification and verification of perception SOTIF entropy in long-tail autonomous driving scenarios.

The performance of the ensemble on subsets of the test dataset PeSOTIF is shown in Table V, where the uncertainty threshold is set to 1.0. The indicators of the 'environment' subset are closer to those of the total dataset, while those of the 'object' subset are much better. Therefore, the algorithm has a certain tolerance for the abnormalities of the appearance and posture of known objects, while the environmental noise has a greater impact on the perception and cognition abilities. The indicators of the 'natural' subset are closer to those of the 'environment' subset, while those of the 'handcraft' subset are much better. No matter the simulation of environmental factors in the experiments, or the adversarial scenarios in the synthetic data sets, the long-tail property of artificial scenarios is not as good as the naturally encountered scenarios.

TABLE. V: Evaluation Results Based on the Protocol

| Metrics | mAP | mAR | mmAP | ACR | FAR | CQS | UQS |
|---|---|---|---|---|---|---|---|
| Total | 0.447 | 0.287 | 0.213 | 0.900 | 0.108 | 0.858 | 4.238 |
| Environment | 0.441 | 0.276 | 0.205 | 0.895 | 0.113 | 0.849 | 3.824 |
| Object | 0.603 | 0.348 | 0.341 | 0.965 | 0.087 | 0.912 | 6.138 |
| Natural | 0.422 | 0.318 | 0.241 | 0.906 | 0.103 | 0.867 | 4.543 |
| Handcraft | 0.630 | 0.317 | 0.275 | 0.899 | 0.114 | 0.840 | 3.007 |

This study uses an NVIDIA Quadro P4000 GPU for the above simulations. As the YOLOv5, deep ensembles, and BSASexcl algorithms are all implemented through Python 3.8 and Pytorch 1.9.0, the perception SOTIF entropy and warnings could be output only at the speed of 20 fps when processing local data. Therefore, the C++ version of the YOLOv5 is developed through TensorRT [41], other algorithms are rewritten into C++ codes, and a multi-threaded calling method is added to deep ensembles to achieve real-time performance. The modified system is deployed to the autonomous driving system of Apollo D-KIT [42], which uses an NVIDIA Geforce RTX2070S GPU. The running speed of the algorithms when processing local videos can reach 100 fps. When acquiring and processing the real-time image data from the camera through CyberRT, the stable running speed is approximately 30 fps.

## V. CONCLUSIONS AND FUTURE WORK

This paper releases a challenging visual dataset PeSOTIF for researches on perception SOTIF problems. The proposed diverse PeSOTIF dataset covers a list of trigger conditions related to visual perception, and the introduced evaluation metrics consider the intuitive feelings of human drivers when verifying whether the system performance is consistent with what human drivers expect. Moreover, as a demonstration, this work proposes a method to quantify and verify the perceptual SOTIF entropy of a 2D probabilistic object detector in the autonomous driving system. In addition, it focuses on the use of the test dataset PeSOTIF and the corresponding evaluation protocol to verify the effectiveness of the proposed method. Specifically, the epistemic uncertainty of the YOLOv5 network is estimated utilizing the deep ensembles and further processed as perceptual SOTIF entropy to quantify the risks.

However, the proposed PeSOTIF dataset is not sufficient for training currently, and many summarized trigger conditions have not been involved and combined yet. In future studies, the authors won't stop enriching the list of analyzed trigger conditions and constructing more perception SOTIF scenarios. Besides, more diversified data, such as LiDAR point clouds, will be collected and preprocessed for multi-modal applications. For the on-board algorithm, the authors will also continue to study how the perception SOTIF entropy can be propagated, utilized, and mitigated in prediction and planning modules downstream in the Apollo autopilot system to achieve safer obstacle detection and avoidance performance.

## ACKNOWLEDGMENT

The authors would like to appreciate the contributions of the perception task group of the CAICV-SOTIF technical alliance in China and the financial support of the National Natural Science Foundation of China Project: U1964203 and 52072215 and the National Key R&D Program of China: 2020YFB1600303.